\documentclass[sigconf]{acmart}
\usepackage{graphicx}
\usepackage{booktabs}
\usepackage{subcaption}
\usepackage{siunitx}
\usepackage{float}

\AtBeginDocument{%
  \providecommand\BibTeX{{%
    \normalfont B\kern-0.5em{\scshape i\kern-0.25em b}\kern-0.8em\TeX}}}

\renewcommand\footnotetextcopyrightpermission[1]{} 
\renewcommand\footnotetextcopyrightpermission[1]{}
\acmConference{}{}{}
\acmBooktitle{}
\acmYear{}
\acmISBN{}
\acmDOI{}
\graphicspath{{figs/}}
\usepackage{xurl} 
\usepackage{hyperref}
\hypersetup{hidelinks}
\begin{document}

\title{Leakage Safe Graph Features for Interpretable Fraud Detection in Temporal Transaction Networks}

\author{Hamideh Khaleghpour}
\email{hamideh-khaleghpour@utulsa.edu}
\orcid{0000-0003-1441-0433}
\affiliation{
  \institution{The University of Tulsa}
  \streetaddress{800 S Tucker Dr}
  \city{Tulsa}
  \state{Oklahoma}
  \postcode{74104}
  \country{USA}
}

\author{Brett McKinney}
\email{brett-mckinney@utulsa.edu}
\affiliation{
  \institution{The University of Tulsa}
  \streetaddress{800 S Tucker Dr}
  \city{Tulsa}
  \state{Oklahoma}
  \postcode{74104}
  \country{USA}
}

\renewcommand{\shortauthors}{Khaleghpour and McKinney}

\begin{abstract}
Illicit transaction detection is often driven by transaction level attributes however, fraudulent behavior may also manifest through network structure such as central hubs, high flow intermediaries, and coordinated neighborhoods. This paper presents a time respecting, leakage safe (causal) graph feature extraction protocol for temporal transaction networks and evaluates its utility for illicit entity classification. Using the Elliptic dataset, we construct directed transaction graphs and compute interpretable structural descriptors, including degree statistics, PageRank, HITS hub or authority scores, $k$-core indices, and neighborhood reachability measures. To prevent look ahead bias, we additionally compute causal variants of graph features using only edges observed up to each timestep. A Random Forest classifier trained with strict temporal splits achieves strong discrimination on a held out future test period (ROC-AUC $\approx 0.85$, Average Precision $\approx 0.54$). Although transaction attributes remain the dominant predictive signal, graph derived features provide complementary interpretability and enable risk context analysis for investigation workflows. We further assess operational utility using Precision at $K$ and evaluate probability reliability via calibration curves and Brier scores, showing that calibrated models yield better aligned probabilities for triage. Overall, the results support causal graph feature extraction as a practical and interpretable augmentation for temporal fraud detection pipelines.
\end{abstract}




\maketitle

\section{Introduction}
Financial fraud and illicit activity increasingly exploit complex transaction ecosystems, where malicious behavior may not be evident from individual transactions in isolation. In real-world payment and cryptocurrency systems, illicit flows often exhibit \emph{network-level} signatures such as highly connected intermediaries, abnormal centrality patterns, coordinated neighborhoods, and atypical multi-hop reachability. These structural signals motivate the use of graph-based representations for fraud detection.

Despite their promise, graph-based methods for temporal fraud detection introduce a critical methodological risk: \emph{look-ahead bias}. In temporal transaction networks, graph features computed on the full graph may inadvertently incorporate edges that occur in the future relative to the prediction time. This leakage can substantially inflate evaluation metrics and produce misleading conclusions about deployable performance. Therefore, ensuring time-respecting feature computation is essential for building trustworthy fraud detection pipelines.

In this work, we propose and evaluate a \emph{causal, time respecting} graph feature extraction protocol for temporal transaction graphs. Our approach computes interpretable structural descriptors for each transaction node while explicitly preventing leakage by restricting feature computation to historical edges available up to the corresponding timestep. We then integrate these graph-derived features with transaction-level attributes in supervised learning models and evaluate performance under a strict time-based split that reflects real deployment, where models must generalize to future unseen periods.

\subsection{Contributions}
The main contributions of this paper are summarized as follows:
\begin{enumerate}
    \item \textbf{Causal graph feature extraction protocol:} We introduce a time-respecting method that prevents look-ahead bias by restricting graph computations to edges observed up to each timestep, enabling leakage-safe evaluation and deployment.
    \item \textbf{Interpretable structural feature suite:} We compute a comprehensive set of graph descriptors including degree statistics, PageRank, HITS hub and authority scores, $k$-core indices, neighborhood degree summaries, and two-hop reachability proxies, along with stabilizing log transforms for heavy-tailed distributions.
    \item \textbf{Operationally grounded evaluation:} Beyond ROC-AUC and Average Precision, we report threshold-based behavior using confusion matrices and investigate ranked alert performance via Precision at $K$, reflecting realistic investigator triage constraints.
    \item \textbf{Probability reliability assessment:} We evaluate probability calibration using calibration curves and Brier scores, and compare calibrated and uncalibrated models to improve the reliability of risk scores for decision support.
\end{enumerate}

\section{Related Work}
Fraud detection has traditionally relied on supervised learning over engineered transaction-level features, including statistical aggregates, temporal indicators, and domain specific heuristics \cite{Ali2022FinancialFraud}. Early graph based approaches extended this paradigm by modeling financial transactions as networks, demonstrating that relational structure and connectivity patterns can be leveraged to identify anomalous and fraudulent behavior \cite{Zhang2016GraphFraudDetection}.

More recent survey studies report strong performance from machine learning and deep learning models for financial fraud detection, but consistently highlight unresolved challenges related to temporal leakage, model interpretability, and deployment realism in operational environments \cite{Sarna2025AIDrivenFraud}. Broader surveys spanning AI, machine learning, and cybersecurity in banking and cryptocurrency systems summarize commonly adopted learning models (e.g., SVM, ANN, and RNN variants) and emphasize security driven research opportunities for fraud and risk analytics in financial networks \cite{Choithani2024AICryptoBankingSurvey}.

Graph based methods, including graph neural networks, have gained increasing attention due to their ability to model relational dependencies and propagation effects in transaction networks \cite{Motie2024GNNFraud}. While these approaches often achieve strong predictive performance, they may incur substantial computational cost and reduce transparency, which can limit their applicability in investigative and regulatory settings. Recent work has also applied GNNs combined with adaptive filtering techniques to detect illicit activity in cryptocurrency transaction graphs, further demonstrating the value of relational learning in decentralized financial systems \cite{Hamidi2026DetectingBitcoin}.

Beyond finance specific studies, broader cybersecurity surveys highlight evolving threat landscapes and attack vectors spanning digital payment systems and financial networks, reinforcing the need for robust, interpretable machine learning and graph based techniques for anomaly and fraud detection in cyber enabled environments \cite{Kaur2022CyberSecurityReview}. In parallel, foundational surveys of blockchain systems have examined the security and privacy properties of public ledgers, emphasizing the structural and adversarial characteristics that underlie fraudulent behavior in cryptocurrency transaction networks \cite{Conti2018BitcoinSurvey}.

Finally, hybrid machine learning frameworks that integrate interpretable models with evolutionary optimization have been explored in other application domains, demonstrating that transparency and robustness can be jointly achieved without sacrificing predictive performance \cite{Khaleghpour2024NeuroFuzzy}.

\subsection{Dataset and Problem Setup}
We evaluate our approach using the Elliptic dataset, which consists of transaction nodes with feature vectors and directed edges representing transaction flows. Each node corresponds to a unique transaction identifier and is associated with an anonymized attribute vector. Labels are provided in three categories: \emph{licit}, \emph{illicit}, and \emph{unknown}. Following standard practice, we train and evaluate models only on labeled instances (licit and illicit) and exclude unknown labels from supervised learning.

\textbf{Temporal evaluation protocol.}
To measure generalization to future time periods, we apply a strict time-based split:
\begin{itemize}
  \item \textbf{Training set:} timesteps $\leq 34$
  \item \textbf{Validation set:} timesteps 35--41
  \item \textbf{Test set:} timesteps $\geq 42$
\end{itemize}
This split reflects a realistic deployment setting in which models are trained on historical data and evaluated on future transactions. We observe class imbalance across all splits, with illicit transactions forming a minority class. Moreover, the illicit rate decreases across time, indicating temporal distribution shift that can affect generalization.

\subsection{Methodology}
\label{sec:methodology}
This section describes the end-to-end pipeline used to construct time-respecting transaction graphs, extract transaction-level and graph-based features, train supervised classifiers, and evaluate predictive performance and operational utility. All experiments follow a strict temporal protocol to prevent look-ahead bias. Classical approaches to fraud detection combine statistical modeling with machine learning techniques to distinguish anomalous behavior from legitimate activity, providing foundational perspectives for modern ML-based detectors \cite{BoltonHand2002}.  

In financial technology and digital banking environments, cybersecurity risk assessment increasingly relies on data-driven analytics pipelines that integrate feature extraction, predictive modeling, and decision support mechanisms to mitigate fraud and systemic risk \cite{Micheal2024FinTechCyberRisk}. Empirical studies in banking and financial regulation further emphasize the importance of robust analytical frameworks for managing cybersecurity risk exposure in operational decision-making contexts \cite{CyberRiskBankLoans2024}.  

Recent methodological surveys highlight how machine learning pipelines are used as core components of risk monitoring, anomaly detection, and threat mitigation strategies across financial infrastructures \cite{WileyCyberAnalytics2023}. In particular, structured ML workflows that combine domain-specific feature engineering with supervised learning have been widely adopted for security analytics in FinTech systems \cite{SpringerFinTechSecurity2024}.  

More broadly, cybersecurity research underscores the need for systematic and auditable analytical pipelines that support risk identification, assessment, and mitigation in complex financial ecosystems \cite{SpringerCyberRiskReview2021}. Such frameworks motivate the design of transparent, modular methodologies that can be deployed in real-world digital banking and payment systems \cite{WileyFinancialSecurity2022}.  

Finally, recent studies in applied information and economic analytics emphasize the growing role of machine learning-based decision pipelines in supporting cybersecurity governance, fraud prevention, and risk-aware financial operations \cite{JIEASCyberRiskAnalytics}.

\subsubsection{Problem Definition}
We study binary illicit transaction classification, where each labeled transaction node is assigned a target label $y \in \{0,1\}$ indicating \emph{licit} (0) or \emph{illicit} (1) behavior. The objective is to learn a function $f(\mathbf{x})$ that predicts the probability of illicit activity from transaction attributes and graph-derived structural context.

\subsubsection{Graph Construction}
\label{sec:graph_construction}
For each timestep $t$, we construct a directed transaction graph $G_t=(V_t,E_t)$, where each node $v \in V_t$ represents a transaction identifier and each directed edge $(u \rightarrow v) \in E_t$ represents a flow relationship between transactions. For features defined on undirected topology (e.g., $k$-core and neighbor degree summaries), we additionally consider an undirected projection of the graph, denoted $G_t^{(u)}$.

\subsubsection{Transaction-Level Attributes}
\label{sec:tx_attributes}
Each transaction node is associated with a fixed-dimensional anonymized feature vector provided by the dataset. These transaction-level attributes serve as the baseline representation and are included in all evaluated models.

\subsubsection{Structural Graph Features}
\label{sec:struct_features}
To incorporate network structure, we compute interpretable graph-derived features for each transaction node, including:
\begin{itemize}
  \item \textbf{Degree statistics:} in-degree, out-degree, and total degree.
  \item \textbf{Centrality measures:} PageRank and HITS hub/authority scores.
  \item \textbf{Cohesiveness:} $k$-core number computed on the undirected projection.
  \item \textbf{Neighborhood context:} mean and maximum neighbor degree, and two-hop reachability proxies.
\end{itemize}
Since transaction graphs often exhibit heavy-tailed distributions, we apply stabilizing transformations where appropriate, including $\log(1+x)$ variants for degree- and reach-based features.

\subsubsection{Causal (Time-Respecting) Feature Extraction}
\label{sec:causal_features}
\textbf{Key novelty.}
To prevent look-ahead bias, we compute causal variants of structural features using only information available up to each timestep. Specifically, for each timestep $t$, we define the historical subgraph:
\[
G_{\leq t} = (V_{\leq t}, E_{\leq t}),
\]
where $E_{\leq t}$ contains only edges observed at or before time $t$. Graph-based features for transactions at timestep $t$ are computed on $G_{\leq t}$ rather than the full graph. This ensures that no future edges influence the feature values, producing leakage-safe graph descriptors suitable for deployment.

\subsubsection{Modeling and Training}
\label{sec:modeling}
We train supervised classifiers using transaction features alone, graph features alone, and hybrid combinations. Given the class imbalance, we report threshold-free metrics including ROC-AUC and Average Precision. Model selection and hyperparameter tuning are performed on the validation set. Threshold-based decision rules are also evaluated using validation-driven threshold selection.

\subsubsection{Operational Evaluation: Thresholding and Ranking}
\label{sec:operational_eval}
To assess practical utility under investigation constraints, we compute confusion matrices at selected thresholds and report precision-recall trade offs across thresholds. Such threshold based and ranking-oriented evaluation strategies are widely adopted in operational cybersecurity and fraud analytics to align predictive models with real-world resource and decision constraints \cite{OperationalMLCyberSecurity2025}. In addition, we evaluate ranked alert performance using \emph{Precision at $K$}, which measures the fraction of true illicit transactions among the top-$K$ highest-risk predictions, reflecting scenarios where investigators can review only a limited number of alerts.

\subsubsection{Probability Calibration}
\label{sec:calibration}
In operational settings, model outputs are often interpreted as probabilities to support risk-based decision making. We evaluate probability reliability using calibration curves and Brier scores. We further compare uncalibrated outputs with calibrated variants, including sigmoid and isotonic calibration, to determine whether calibrated probabilities better align with empirical outcome frequencies and improve interpretability for triage workflows.

\section{Models and Evaluation}
\label{sec:models_evaluation}

This section presents the predictive models and evaluation methodology used to quantify the effectiveness of time respecting structural graph descriptors for illicit transaction detection. Our experimental design is constructed to reflect realistic deployment conditions in which models are trained on historical data and must generalize to future transaction behavior, while avoiding information leakage from future graph edges.

\subsection{Predictive Task}
\label{sec:task}

Let each transaction be represented by a feature vector $\mathbf{x}_i$ and a binary label $y_i \in \{0,1\}$, where $y_i=1$ denotes an illicit transaction and $y_i=0$ denotes a licit transaction. The objective is to learn a scoring function $f(\mathbf{x})$ that assigns a risk score $s_i=f(\mathbf{x}_i)$, which is subsequently used for ranking, threshold-based classification, and operational triage.

\subsection{Feature Configurations}
\label{sec:feature_configs}

To isolate the contribution of network structure relative to transaction-level attributes, we evaluate the following feature configurations:

\begin{itemize}
    \item \textbf{Transaction-only (T):} Uses only the original transaction attributes provided in the dataset.
    \item \textbf{Graph-only (G):} Uses only structural graph descriptors computed from the transaction network, including centrality, degree-based measures, neighborhood statistics, and their stabilized variants.
    \item \textbf{Hybrid (T+G):} Combines transaction attributes and structural graph descriptors into a unified representation.
\end{itemize}

This design enables a direct assessment of whether graph structure provides complementary predictive signal beyond the transaction-level baseline.

\subsection{Learning Model}
\label{sec:model}

We employ a Random Forest classifier as the primary learning algorithm. Random Forests provide strong performance on heterogeneous tabular features, naturally capture non-linear feature interactions, and offer transparency through feature importance analysis, which is valuable for high-stakes fraud detection settings where interpretability is often required.

Given training samples $\{(\mathbf{x}_i,y_i)\}_{i=1}^{n}$, the classifier estimates an illicit probability $\hat{p}_i = P(y_i=1\mid \mathbf{x}_i)$ for each transaction. We use class-weighted training to account for label imbalance, ensuring that illicit instances contribute proportionally more to the loss optimization.

\subsection{Temporal Evaluation Protocol}
\label{sec:temporal_protocol}

Performance is evaluated using a strict chronological split based on transaction timesteps:
\begin{itemize}
  \item \textbf{Training period:} timesteps $\leq 34$
  \item \textbf{Validation period:} timesteps 35--41
  \item \textbf{Test period:} timesteps $\geq 42$
\end{itemize}

This protocol reflects a forward-in-time evaluation setting. The validation set is used exclusively for model selection, threshold selection, and calibration fitting. The test set is held out for final reporting to preserve the integrity of the evaluation.

\subsection{Evaluation Metrics}
\label{sec:metrics}

Illicit transaction detection is characterized by severe class imbalance, making accuracy insufficient and potentially misleading. We therefore report complementary metrics that reflect ranking quality, rare-event detection performance, and decision making behavior under operational constraints.

\subsubsection{Discrimination Metrics}
\label{sec:discrimination}

We report:
\begin{itemize}
    \item \textbf{ROC-AUC:} Measures the ability of the model to rank illicit transactions above licit ones across all possible thresholds.
    \item \textbf{Average Precision (AP):} Summarizes the precision recall curve and is particularly informative when the positive class is rare.
\end{itemize}

\subsubsection{Threshold Based Decision Analysis}
\label{sec:threshold_eval}

To translate risk scores into actionable decisions, we analyze performance at explicit decision thresholds. We compute confusion matrices and derive standard classification statistics (precision, recall, and F1 score). Thresholds are determined using the validation set to avoid optimistic bias on the test set. In addition to reporting results at a conventional threshold (e.g., $0.5$), we include validation optimized thresholds that reflect different operational priorities, such as high recall screening or precision focused alerting.

\subsubsection{Top-$K$ Alert Precision}
\label{sec:topk}

In many investigation workflows, analysts review only a limited number of highest risk alerts. To reflect this setting, we evaluate ranked alert utility using \textbf{Precision at $K$}, defined as the proportion of illicit transactions among the top-$K$ highest scoring predictions. This metric captures the quality of the prioritized investigation queue and directly reflects resource constrained triage effectiveness.

\subsection{Probability Reliability and Calibration}
\label{sec:prob_calibration}

While discrimination metrics assess ranking quality, operational systems often require probability estimates that are meaningful as risk measures. In particular, investigators and downstream automation may interpret $\hat{p}_i$ as an estimate of the true likelihood of illicit behavior. However, ensemble classifiers such as Random Forests are frequently miscalibrated, producing probabilities that are systematically over or under confident.

We assess probability reliability using:
\begin{itemize}
    \item \textbf{Calibration curves:} Compare predicted probabilities against empirical outcome frequencies across probability bins.
    \item \textbf{Brier score:} Measures the mean squared error between predicted probabilities and observed binary outcomes, where lower values indicate better calibration.
\end{itemize}

To improve probability alignment, we apply post hoc calibration methods trained on the validation set, including sigmoid calibration and isotonic regression. Calibrated models are evaluated on the test period to determine whether probability reliability improves without materially degrading discrimination performance.

\subsection{Experimental Reporting and Reproducibility}
\label{sec:reproducibility}

All results are reported on the held out test period using the same feature extraction and temporal split protocol. To support reproducibility, we fix random seeds where applicable and maintain consistent feature column ordering across train, validation, test matrices. Missing graph derived values (e.g., for nodes absent from a given historical subgraph) are imputed with zeros, and heavy tailed structural measures are stabilized using $\log(1+x)$ transforms.

\section{Results and Discussion}
\label{sec:results}

This section reports predictive performance on a held-out future test period and analyzes the role of graph-derived descriptors under a strict time-respecting (causal) feature extraction protocol. In addition to threshold-free discrimination metrics, we evaluate operational behavior via threshold-based confusion matrices, ranked alert precision, and probability reliability for triage.

\subsection{Temporal Split and Class Imbalance}
\label{sec:results_split}

Table~\ref{tab:split_stats} summarizes the labeled class distribution across the temporal split. The dataset is highly imbalanced, and the illicit rate decreases in later timesteps, indicating temporal distribution shift. This forward-in-time evaluation is intentionally challenging and better reflects deployment conditions, where fraud patterns and transaction network dynamics evolve over time.

\begin{table}[t]
\centering
\caption{Temporal split statistics (labeled transactions only).}
\label{tab:split_stats}
\begin{tabular}{lrr}
\toprule
\textbf{Split} & \textbf{Licit (0)} & \textbf{Illicit (1)} \\
\midrule
Train ($t \leq 34$) & 26,432 & 3,462 \\
Validation ($35 \leq t \leq 41$) & 7,154 & 675 \\
Test ($t \geq 42$) & 8,433 & 408 \\
\bottomrule
\end{tabular}
\end{table}

\subsection{Overall Discrimination Performance}
\label{sec:results_main}

We first evaluate the hybrid Random Forest model that combines transaction attributes with causal (time-respecting) graph descriptors. Table~\ref{tab:rf_main} reports ROC-AUC and Average Precision (AP) on validation and test sets. As expected under temporal shift, performance decreases from validation (ROC-AUC $=0.977$, AP $=0.925$) to the held-out future test period (ROC-AUC $=0.853$, AP $=0.537$). Importantly, the test-period results remain meaningfully above chance, indicating that the model retains useful ranking capability when applied to future timesteps.

\begin{table}[t]
\centering
\caption{Random Forest performance on validation and future test sets.}
\label{tab:rf_main}
\begin{tabular}{lcc}
\toprule
\textbf{Split} & \textbf{ROC-AUC} & \textbf{AP} \\
\midrule
Validation & 0.977 & 0.925 \\
Test & 0.853 & 0.537 \\
\bottomrule
\end{tabular}
\end{table}

\begin{figure}[t]
  
  \centering
  \includegraphics[width=0.85\linewidth]{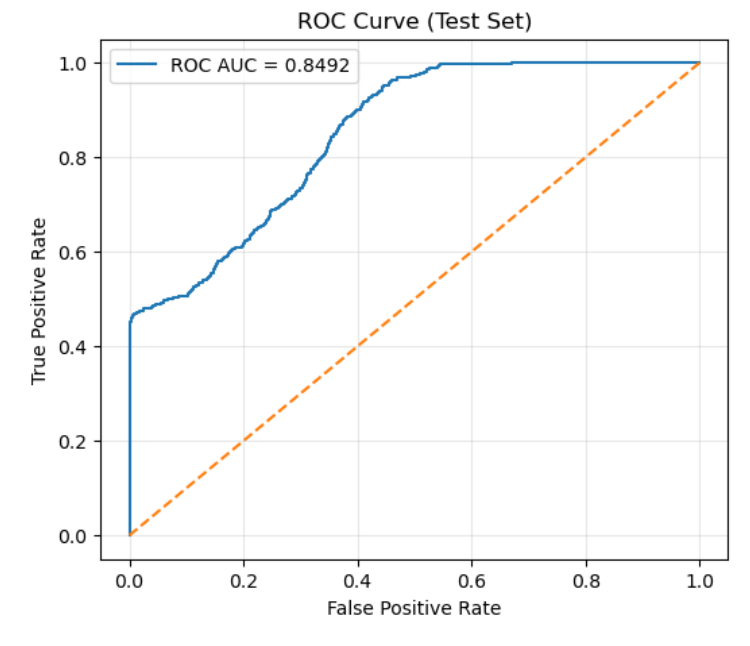}
    \caption{ROC curve on the test set.}
 \Description{Receiver operating characteristic curve for the hybrid Random Forest model on the held-out test period, demonstrating discrimination performance across thresholds.}
  \label{fig:roc_test}
 \end{figure}

Fig.~\ref{fig:roc_test} visualizes discrimination on the test set using the ROC curve. The ROC curve summarizes ranking quality across thresholds, while the PR curve is particularly informative under class imbalance and highlights performance in the high risk region where investigative attention is typically concentrated.

\subsection{Ablation Study: Transaction vs.\ Graph Features}
\label{sec:results_ablation}

To isolate the contribution of structural descriptors, we compare transaction-only (T), graph-only (G), and hybrid (T+G) feature sets. Table~\ref{tab:ablation} shows that transaction attributes dominate predictive performance on Elliptic: the transaction-only model (ROC-AUC $=0.847$, AP $=0.537$) is essentially matched by the hybrid model (ROC-AUC $=0.853$, AP $=0.537$), while the graph-only model performs substantially worse (ROC-AUC $=0.562$, AP $=0.049$).

These results suggest that, for this benchmark, handcrafted structural descriptors provide limited incremental discrimination beyond transaction-level attributes. Nevertheless, graph-derived features can still be operationally valuable by providing interpretable network context for flagged transactions (e.g., unusually central nodes, hub/authority structure, or dense neighborhoods), supporting analyst reasoning even when aggregate AUC/AP gains are modest.

\begin{table}[t]
\centering
\caption{Feature ablation study on the test set.}
\label{tab:ablation}
\begin{tabular}{lcc}
\toprule
\textbf{Feature Set} & \textbf{ROC-AUC} & \textbf{AP} \\
\midrule
Transaction-only (T) & 0.847 & 0.537 \\
Graph-only (G) & 0.562 & 0.049 \\
Hybrid (T+G) & 0.853 & 0.537 \\
\bottomrule
\end{tabular}
\end{table}

\subsection{Threshold Behavior and Operational Triage}
\label{sec:results_threshold}

\begin{figure}
    \centering
    \includegraphics[width=0.85\linewidth]{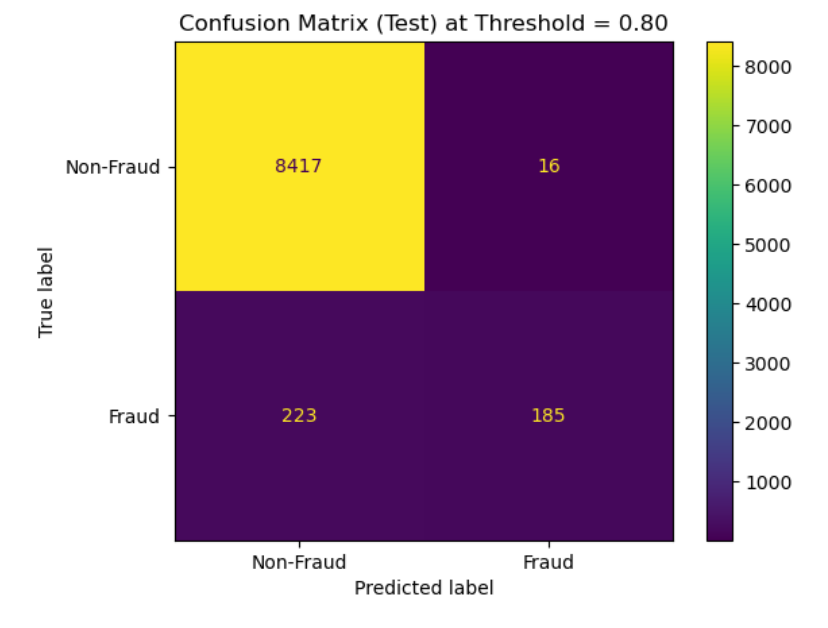}
     \caption{Test-period confusion matrix at probability threshold $=0.80$.}
  \Description{Confusion matrix on the held-out test set at a precision-focused operating threshold of 0.80, illustrating the trade-off between false positives and false negatives.}
  \label{fig:cm_thr080}
\end{figure}

In operational deployments, risk scores must often be converted into actions (e.g., alert creation) under limited investigative capacity. Fig.~\ref{fig:cm_thr080} reports the test-period confusion matrix at a representative operating threshold. As the decision threshold increases, false positives decrease (reducing analyst workload), while false negatives increase (missing more illicit cases). This trade off highlights that threshold selection should be driven by operational constraints and risk tolerance, rather than by a fixed cutoff applied universally across time.

\begin{figure}[t]
    \centering
    \includegraphics[width=0.85\linewidth]{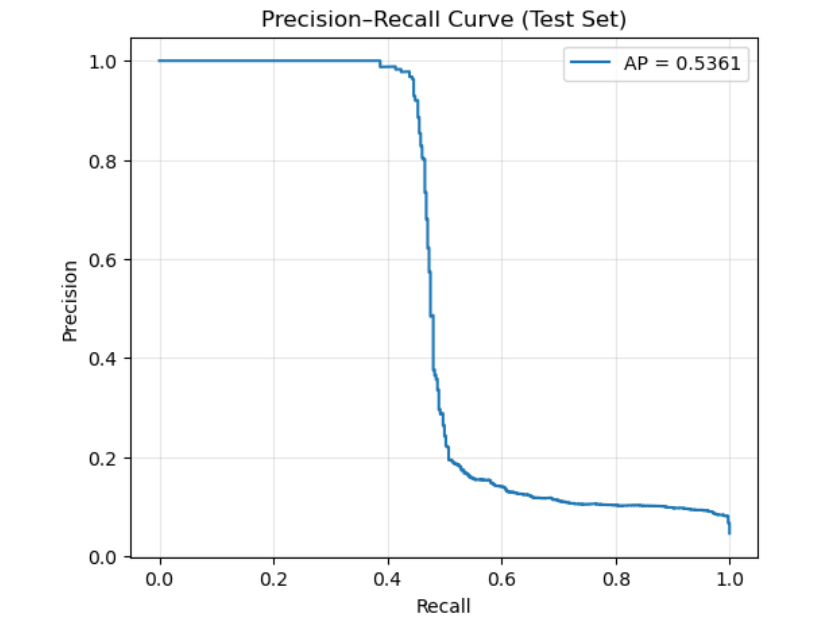}
    \caption{Precision--Recall curve on the test set.}
    \Description{Precision-Recall curve for the hybrid Random Forest model on the held-out test set, summarizing the trade off between precision and recall under class imbalance.}
    \label{fig:pr_test}
\end{figure}

In addition to threshold-based classification, we evaluate ranking based triage using Precision-Recall (PR) analysis, which assesses how illicit transactions are prioritized as the alert budget increases. Fig.~\ref{fig:pr_test} shows the PR curve on the test set, summarizing the precision-recall trade off under severe class imbalance.

\begin{figure}
    \centering
    \includegraphics[width=0.85\linewidth]{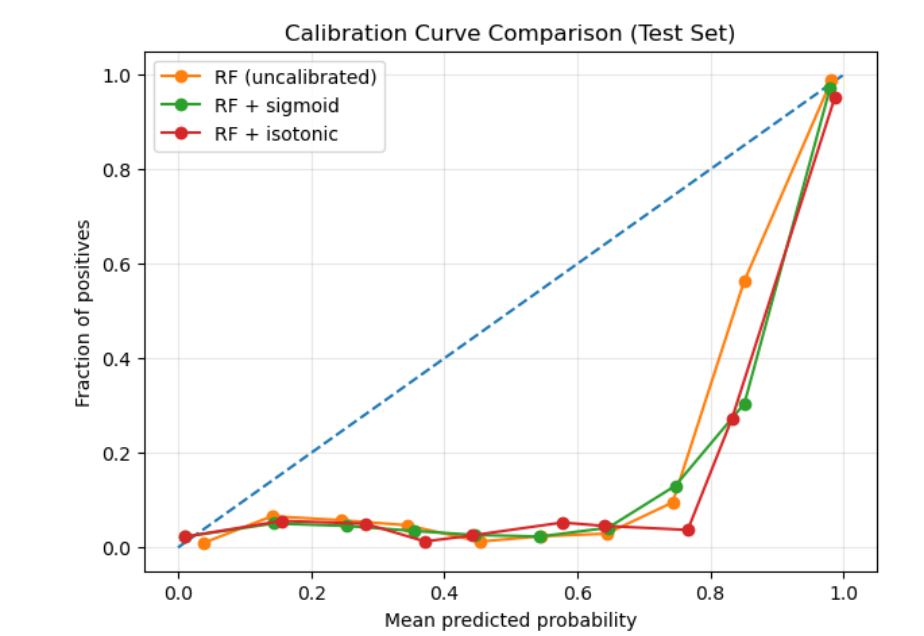}
\caption{Calibration curve comparison on the test set for the Random Forest model.}
\Description{Reliability diagrams comparing uncalibrated Random Forest probabilities with sigmoid- and isotonic-calibrated outputs on the held-out test set. The diagonal line represents perfect calibration.}
\label{fig:calibration_compare}
\end{figure}

\begin{figure}
    \centering
    \includegraphics[width=0.85\linewidth]{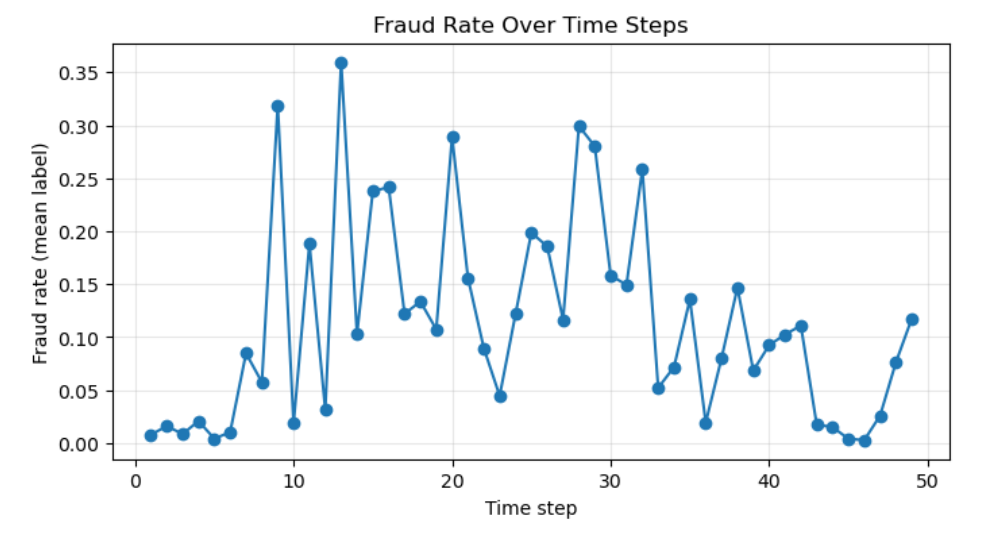}
    \caption{Temporal evolution of the illicit transaction rate across timesteps.}
\Description{Mean illicit label rate per timestep, illustrating class imbalance and temporal distribution shift in the dataset.}
\label{fig:fraud_rate_time}
\end{figure}

\begin{figure}
    \centering
    \includegraphics[width=0.85\linewidth]{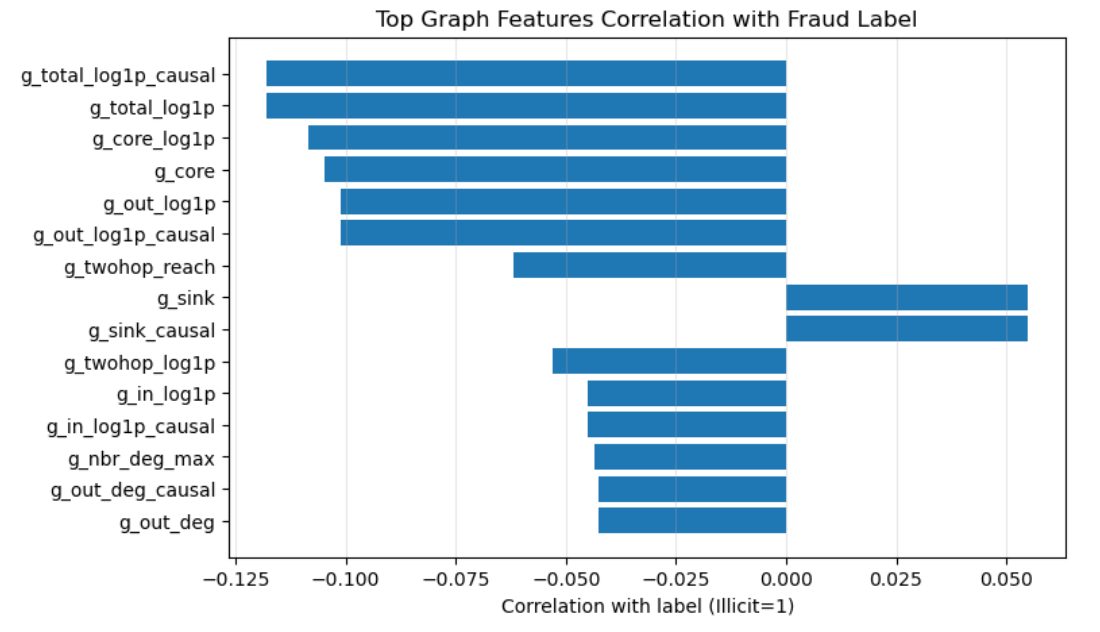}
    \caption{Correlation between structural graph features and the illicit transaction label.}
    \Description{Pearson correlation coefficients between selected graph-derived features and the binary fraud label, highlighting associations between network structure and illicit behavior.}
    \label{fig:graph_corr}
\end{figure}

\subsection{Summary of Findings}
\label{sec:results_summary}

Overall, the results demonstrate that strict time-respecting feature computation yields realistic future-period performance without look-ahead leakage. Transaction attributes provide the dominant predictive signal on Elliptic, while causal structural descriptors contribute interpretable network context that can support analyst understanding and investigation. Finally, probability calibration improves the reliability of risk estimates, which is critical when scores are consumed as probabilities for triage decisions and downstream policy rules.

\begin{figure}
    \centering
    \includegraphics[width=0.85\linewidth]{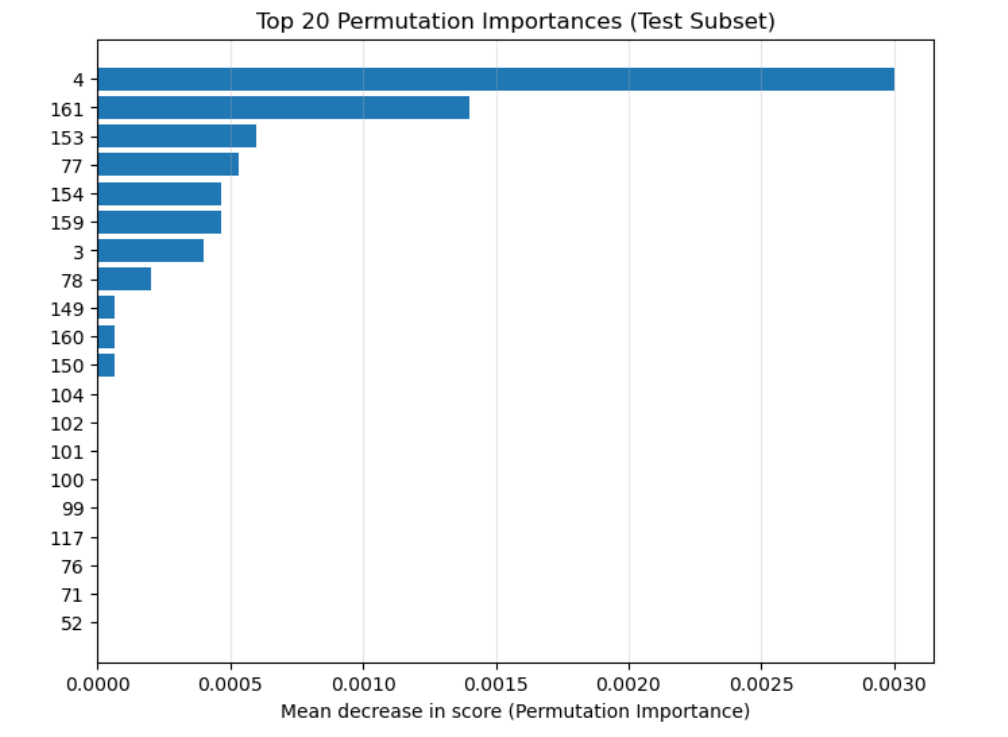}
    \caption{Top permutation feature importances on the test subset.}
    \Description{Permutation importance scores for the top-ranked features, measured as the mean decrease in model performance when each feature is randomly permuted.}
  \label{fig:perm_importance}
\end{figure}

\subsection{Probability Reliability and Calibration}
\label{sec:results_calibration}

For decision support, predicted risk scores are often interpreted as probabilities. However, tree ensembles such as Random Forests are frequently miscalibrated, especially under class imbalance and distribution shift. Fig.~\ref{fig:calibration_compare} presents reliability diagrams for uncalibrated and post hoc calibrated probability estimates, demonstrating improved alignment between predicted probabilities and empirical outcome frequencies after calibration. The calibrated model exhibits improved agreement between predicted probability bins and empirical outcome frequencies, which is important when downstream policies or analyst decisions rely on probability values rather than only on ranking.

\begin{figure}
    \centering
    \includegraphics[width=0.85\linewidth]{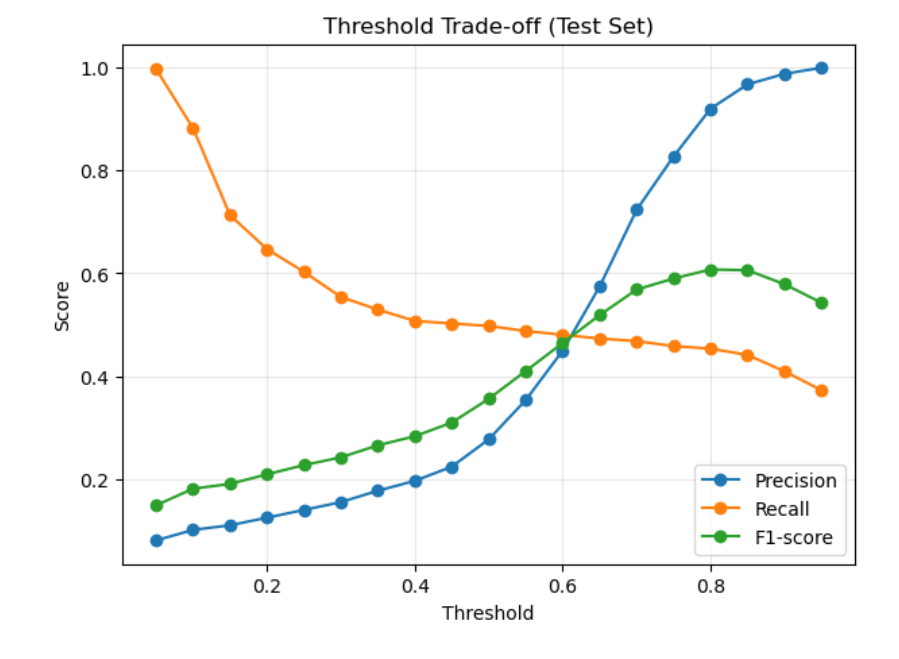}
    \caption{Threshold trade-off analysis on the test set.}
    \Description{Precision, recall, and F1-score as functions of the decision threshold, illustrating operational trade-offs between detection coverage and alert precision.}
    \label{fig:threshold_tradeoff}
\end{figure}

\section{Conclusion and Future Work}
\label{sec:conclusion}

This paper presented a leakage safe, time respecting graph feature extraction protocol for temporal transaction networks and evaluated its utility for illicit transaction classification on the Elliptic dataset. By computing structural descriptors on historical subgraphs up to each timestep, the proposed method prevents look ahead bias and better reflects realistic deployment conditions. A Random Forest classifier trained with strict temporal splits achieved meaningful generalization to future timesteps (test ROC-AUC $\approx 0.85$, AP $\approx 0.54$), demonstrating that the pipeline can provide actionable ranking performance under temporal distribution shift.

While transaction level attributes remain the dominant predictive signal in this benchmark, graph-derived descriptors contribute interpretable structural context, enabling analysts to reason about network position, neighborhood reachability, and centrality patterns associated with high risk transactions. Furthermore, probability calibration improves reliability of predicted risk scores, supporting triage workflows where probabilities are consumed directly.

Future work will explore richer temporal and relational modeling approaches, including (i) more expressive graph representations such as temporal graph neural networks, (ii) feature learning strategies that capture higher order coordinated substructures beyond handcrafted descriptors, (iii) domain adaptive training to mitigate temporal distribution shift, and (iv) evaluation under realistic analyst budgets using cost sensitive metrics and decision theoretic threshold selection. These directions may further improve both predictive accuracy and operational utility in real world financial fraud detection systems.

\bibliographystyle{ACM-Reference-Format}
\nocite{Khaleghpour2024NeuroFuzzy,Khaleghpour2025UnifiedAI}
\bibliography{references}
\end{document}